# Voice-Assisted Real-Time Traffic Sign Recognition System Using Convolutional Neural Network

Mayura Manawadu
*Department of Computer Engineering*
*University of Sri Jayewardenepura*
Nugegoda, Sri Lanka
en86191@sjp.ac.lk

Udaya Wijenayake
*Department of Computer Engineering*
*University of Sri Jayewardenepura*
Nugegoda, Sri Lanka
udayaw@sjp.ac.lk

*Abstract*—Traffic signs are important in communicating information to drivers. Thus, comprehension of traffic signs is essential for road safety and ignorance may result in road accidents. Traffic sign detection has been a research spotlight over the past few decades. Real-time and accurate detections are the preliminaries of robust traffic sign detection system which is yet to be achieved. This study presents a voice-assisted real-time traffic sign recognition system which is capable of assisting drivers. This system functions under two subsystems. Initially, the detection and recognition of the traffic signs are carried out using a trained Convolutional Neural Network (CNN). After recognizing the specific traffic sign, it is narrated to the driver as a voice message using a text-to-speech engine. An efficient CNN model for a benchmark dataset is developed for real-time detection and recognition using Deep Learning techniques. The advantage of this system is that even if the driver misses a traffic sign, or does not look at the traffic sign, or is unable to comprehend the sign, the system detects it and narrates it to the driver. A system of this type is also important in the development of autonomous vehicles.

## I. INTRODUCTION

Since automobiles have become an indispensable medium of transportation, assurance of safety has been implemented in every country through proper road rules and regulations. Among them, traffic signs provide valuable information to the drivers and help to communicate the rules to be followed in that specific area. The purpose of a traffic sign is to convey a message quickly and accurately with minimum reading skills. Negligence, lack of attention, lack of familiarity, accidentally or deliberately not noticing traffic signs, distracting driving behaviours have been discovered as major reasons for the ignorance of road signs among the drivers which eventually lead to road accidents. Furthermore, drivers in unurbanized communities may find it difficult in decoding the message conveyed by a specific road sign due to a lack of familiarity with the plenty of road signs in urbanized areas. Some drivers tend to ignore certain traffic signs believing that they are not necessary. Different attitudes of the drivers are also a reason for this ignorance. Ignorance or unfamiliarity with traffic signs could result in severe accidents and may even cost lives.

To address the above problems, this paper provides a method to detect and recognize traffic signs in real-time with higher accuracy and narrating the signs to the drivers. A system of this type can be used in both vehicle assistive systems and autonomous vehicles. The system is implemented using the Convolutional Neural Network (CNN) model architecture of YOLO [1]. With the faster detection rates and optimized accuracy of the model, the system can be used as a real-time traffic sign detection system. The narration of the message given by a particular traffic sign can assist the drivers while driving. By the voice narration, the issues like missing the traffic signs, lack of familiarity, and the complexity of the traffic signs can be solved.

## II. RELATED WORK

Over the last few decades autonomous driving and assisted driving have become the research spotlight among the research community. Traffic sign detection and recognition have become an important research topic out of these. In traffic sign detection, three common methods are being practised: colour-based [2], shape-based [3], and machine learning-based methods. HSV (Hue, Saturation and Value) transformation can be considered as one of the popular colour-based detection approaches [4]. Even though colour-based techniques are computationally efficient, due to low light conditions, weather changes, or illuminated environments and similar backgrounds, the accuracy becomes low. The shape-based detections [5] extract the shapes like triangles, rectangles, and circles where the traffic signs are likely to be, using techniques like Hough transform, and Edges with Harr-like features. However, this approach is not robust when shapes similar to traffic signs appear. Thus, the limitations of these approaches have led the research community to focus the spotlight on deep learning techniques. Recently, the use of deep CNN [6] has been able to yield robust, fast, and accurate results. Convolutional neural networks can learn features using a large number of training examples without preprocessing, which avoids the difficulties of traditional approaches. The evolution of high-end graphical processing units (GPU) has catalyzed the deep learning approaches and resulted in the development of different types of convolutional neural network architectures such as R-CNN (Region-based Convolutional Neural Networks) [7], YOLO (You Only Look Once) [1] and SSD (Single Shot Multibox Detection) [8].

Generally, the deep learning approaches can be divided into two different methods as two-stage detectors and one-stage detectors [9]. The two-stage detectors are based on the region of interests (ROI). Here, the first stage model proposes a set of ROIs, and the second stage classifies the proposed regions into the candidate classes. R-CNN can be considered as a cutting-edge architecture under two-stage detectors. R-CNN generates ROIs by selective search and extracts the features of CNN separately. At the last stage of the networks, it uses Support Vector Machine (SVM) classifier to predict the classes of objects. To optimize the performance, it uses linear regression to fine-tune the position and sizes of bounding boxes. Other CNN architectures such as spatial pyramid pooling network (SPP-NET) [10] and Faster R-CNN [11] have been evolved as two-stage detectors. The frame rate that was able to reach using two-stage detectors were considerably low.

The other approach, one stage-detector skips the ROI proposal stage and runs as an end-to-end learning model by iterating over the entire image and proposing the prediction values of the candidates at every location of the images. YOLO integrates the object detection and recognition into a single CNN which results in remarkably increased speed than two-stage detectors [1]. SSD is also a type of single-stage detector.

The two-stage networks yield higher accuracy, however the detection rates are slow. On the other hand, one stage networks are faster since it predicts the candidate classes at one stage through the network. But their precision is low when compared with two-stage detectors.

Although there are deep neural network approaches which have been adapted on traffic sign detection and recognition [12], this paper presents a novel method of a one-stage CNN approach based on YOLO architecture along with a voice assistive message to narrate the detected sign to the driver.

## III. METHODOLOGY

### A. Convolutional Neural Networks

CNN is the state-of-the-art deep learning technique used in computer vision. Neural Network is a mathematical model which is modelled based on the primitives of neurons. A large number of artificial neurons are networked into layers to build a deep neural network. It accepts vectors as inputs and passes through the layers of the network and predicts the output. CNN is a type of deep neural network which consists of three types of layers namely convolution, pooling, and fully connected layers. Out of these, the first two types are involved with the extraction of characteristics while the fully connected layers map the extracted features into classification. Several convolutional neural network architectures have been adapted in the process of image detection and image recognition. The CNN that is used in this paper is the YOLO which was proposed by Joseph Redmon *et al* [1].

### B. YOLO Architecture

Several versions of YOLO architectures have been incorporated in the training and testing phase in this paper. Here the focus is set on YOLOv4 which yielded the best results. YOLOv4 was developed by Alexy Bochkovskiy et al. [13] Instead of selecting the Regions of Interest (ROI) as in two-stage detectors like RCNN, the YOLO algorithm [1] predicts classes and bounding boxes from the whole image in just one run in the network. YOLOv4 outperforms the other members in the YOLO family with an Average Precision of 43.5% on COCO dataset with 65 FPS in a Tesla V100. Also, YOLOv4 addresses the need for multiple GPUs by employing an object detector which can be trained on a single GPU. The top-level architecture of YOLOv4 is shown in Fig. 1, which is extracted from the YOLOv4 paper [13].

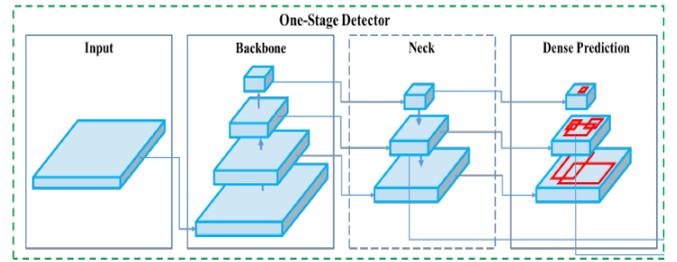

Fig. 1. The high-level architecture of YOLOv4 network [13].

As shown in Fig. 1 the backbone is used as a feature extractor. Authors have set focus on CSPResNext50, CSPDarknet53 and EfficientNet-b3 as backbones for the YOLOv4 object detector. Based on the experiments on ImageNet and MS-COCO datasets, CSPDarknet53 is taken as the backbone for the YOLOv4 detector. The Neck is the set of extra layers that connects the backbone with the head. Neck layers are used to extract different feature mappings at different levels of the backbone. YOLOv4 uses Spatial Pyramid Pooling (SPP) and a modified version Path Aggregation Network (PAN) for the detector. The head part or the dense prediction is the network which is used to carry out the detection parts. Specifically, it carries out the detection and regression of the bounding boxes. YOLOv4 uses the same head as in YOLOv3 [14]. As in Fig. 1, the detections are carried out at 3 YOLO layers. But, for this research, the tiny version of YOLOv4 architecture is used where the network size is dramatically reduced. YOLOv4 architecture uses only 2 YOLO layers and the convolutional layers in the CSP backbone are compressed. Thus, it makes the detections at faster rates.

The ultimate target of the YOLO detection layer is to predict the class of an object and locate the bounding box related to that image. As shown in Fig. 2 the bounding box has four parameters describing it: centre coordinates represented by $(b_x, b_y)$, width $(b_w)$ and height $(b_h)$.

As mentioned earlier, there is no selection of ROI like in two-stage detectors. Instead, the input image is split into $S \times S$ squares. Each square in the grid predicts $B$ number of bounding boxes and their confidence values along with the classes $C$. Confidence values measure whether the square is consisting of objects and if there is any object the accuracy of the bounding box is predicted.

$$Confidence = pr(Object) \times IoU \qquad (1)$$

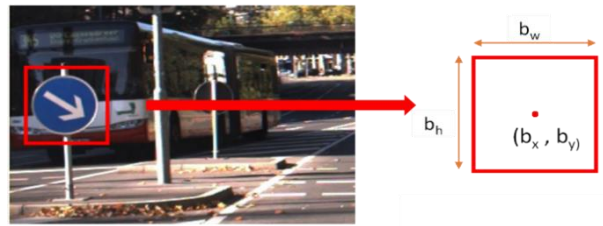

Fig. 2. Format of the bounding box in YOLO network.

When a square in the grid contains a part of the ground truth box of an object, the value of $pr(Object)$ becomes 1 and 0 if there is no ground truth box. IoU (Intersection over Union) indicates the intersection over union values between the predicted bounding box, and the ground truth box. As each bounding box is represented by five values: $b_x, b_y, b_w, b_h$, and confidence, the output is a tensor of shape $S \times S$ ($5 \times B + C$). When multiple frames are predicting the same object, YOLO uses a non-maximum suppression technique to select the most suitable frame.

### C. Audio Narration

The system is integrated with an audio feedback system to narrate the detected traffic signs. As the detections are happening real time, the audio outputs should also be provided in real time. For this, the detections and audio outputs are allowed to run parallelly in such a way that when an object is detected, the voice feedback of the particular sign is provided simultaneously. The gTTS (Google Text To Speech) library is used for audio narrations. The language can be customized as per user preference. Whenever a new detection is observed in the frame, the detected sign is fed into the algorithm where the particular voice is played. Fig. 3 illustrates the high-level architecture including the audio feedback system.

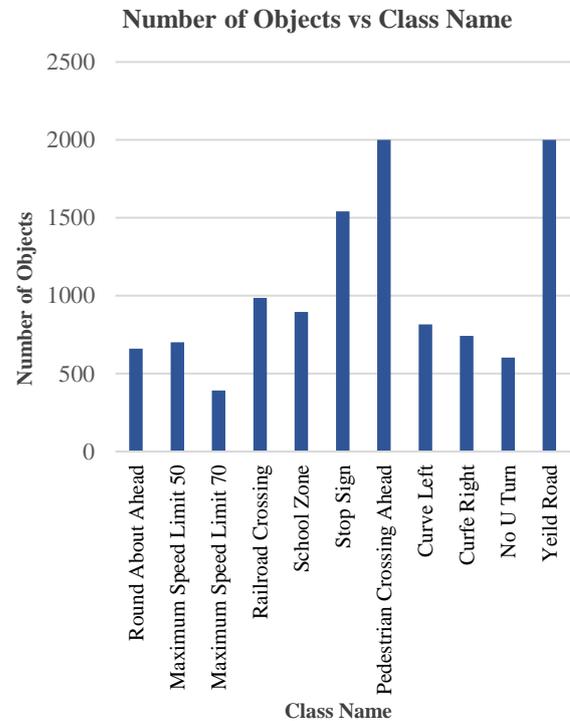

Fig. 3. The high-level architecture of the system.

### D. Dataset

Initially, the system was developed using the German Traffic Sign Detection Benchmark (GTSDB) dataset for the CNN model. The dataset contains 900 images in which 600 are training images and 300 are validation images. Zero to six traffic signs are included per image. The traffic signs in this dataset appear in every perspective and under many lighting conditions. Few example images are given in Fig. 4. The traffic sign instances are divided into four categories as danger, prohibitory, mandatory, and others. The dataset contains annotations in CSV format, and it is converted to YOLO format by developing an algorithm using python. LabelImg tool was used to test the annotations which were converted to YOLO format.

The subsequent models were tested using the images extracted from the Mapillary Traffic Sign Dataset which FPScontains over 100,000 high-resolution images. There are nearly 300 classes of traffic signs covering almost all the continents. The dataset contains images varying under different environmental conditions like rain, sun, snow, dawns, daylight, night etc. Few example images from the dataset are given in Fig. 5. Due to a large number of data and classes, only selected classes were used for the experiment to cope with the system configurations given in TABLE 1. As the dataset is annotated into 300 classes, the number of images available for some classes were not enough for the training purpose. Thus, the classes were integrated in such a way that both regulatory type and warning type of that particular sign is grouped to represent the same sign. Since the annotations were done in JSON format, a separate algorithm was developed to select the desired classes and to convert the JSON annotations into YOLO format. LabelImg tool was used to confirm the correctness of the conversion. Fig. 6 Shows the distribution of a filtered dataset.

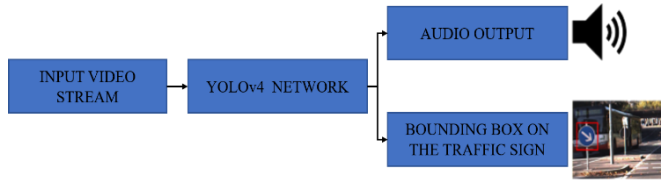

Fig. 4. Sample images from GTSDB.

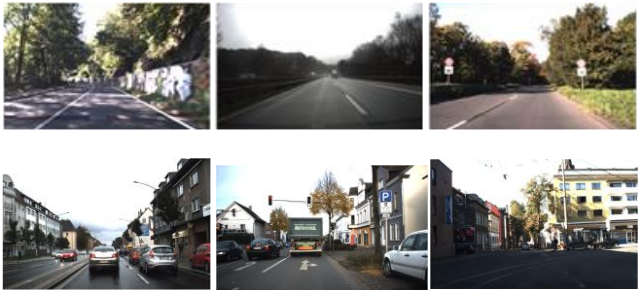

Fig. 5. Sample images from Mapillary Traffic Sign Dataset.

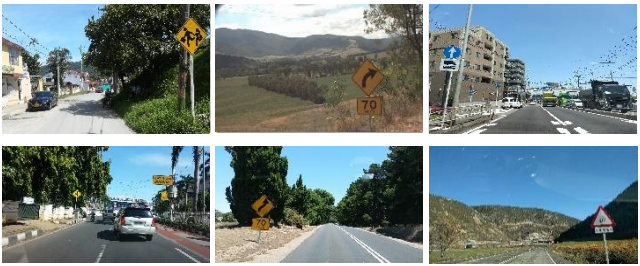

**Number of Objects vs Class Name**

Fig. 6. Dataset Distribution.

TABLE I.  System and environment configurations

| Hardware/Software | Specification |
|---|---|
| CPU | Intel® Core i7-7700HQ |
| GPU | Nvidia GTX 1050 |
| RAM | 8GB |
| Operating system | Ubuntu 18.04 |
| CUDA Version | 10.0 |

TABLE II.  Summary of the Outputs from the trained models

| Model Number | Detector | # Classes | mAP @ 50 (%) | Precision | Recall | F1 Score | Average FPS |
|---|---|---|---|---|---|---|---|
| 1 | YOLOv3 | 4 | 89.75 | 0.98 | 0.84 | 0.9 | 16.5 |
| 2 | YOLOv3 Tiny | 16 | 8.11 | 0.15 | 0.1 | 0.12 | 68 |
| 3 | YOLOv4 Tiny | 16 | 46.10 | 0.64 | 0.44 | 0.52 | 58 |
| 4 | YOLOv4 Tiny | 16 | 48.16 | 0.69 | 0.45 | 0.55 | 63 |
| 5 | YOLOv4 | 11 | 51.00 | 0.58 | 0.53 | 0.56 | 32 |
| 6 | YOLOv4 Tiny | 11 | 64.71 | 0.79 | 0.62 | 0.69 | 55 |

## IV. Experimental Results

Different versions of YOLO networks have been trained and tested. TABLE II provides details of the training results.

Initially, the model was trained on YOLOv3 using the Darknet framework. The first model was trained using the GSTDB dataset and the rest of the models were trained using Mapillary Traffic Sign Dataset. The number of classes under selected from the Mapillary Traffic Sign Dataset has been varied to find a training set which is equally distributed among the given classes and to minimize the complexities in detections of the selected classes.

As shown in TABLE II, model number 1 has the highest mAP (Mean Average Precision). But the FPS (Frames Per Second) value is extremely low, which is not eligible for a real-time detection system. Also due to the constrains of only four classes being available in GSTDB dataset the rest of the models were trained using the Mapillary Traffic Sign Dataset. Due to the lower FPS rate, the initial target was set upon in increasing the FPS value, thereafter, increasing the accuracy. Thus, the faster version of YOLO network, tiny YOLO was used. To accelerate the detection speed, OPENCV was integrated with CUDA to run the detections on top of the GPU. Model 2 was developed by applying those. Here it can be seen that the FPS rate which was low in model 1 has been increased dramatically. But on the other hand, the accuracy of the model has been decreased. The next set of models (model number 3,4,5,6) was developed using the YOLOv4 and YOLOv4-tiny which yields robust results than the previous versions at faster rates.

TABLE III.  Summary of the Most Optimum Model (Model 6)

| Dataset Used | Mapillary Traffic Sign Dataset |
|---|---|
| Dataset Size | 11335 Images |
| Number of Classes | 11 |
| Precision | 0.79 |
| Recall | 0.62 |
| F1 Score | 0.69 |
| mAP @ 50 | 64.71% |
| True Positives | 1064 |
| False Positives | 286 |
| False Negatives | 657 |
| Number of Iterations | 22000 |
| Training Time | 22 hrs |
| Input Resolution | 608 * 608 |

Model number 5 was trained using the YOLOv4 and its frame rate was checked on halfway using the weight files at checkpoints. The Model 6 yields outstanding results than the previous models. It was trained using the Tiny YOLOv4 model while adjusting the set of hyperparameters and other configurations of the model. The TABLE III shows the detailed results of the mentioned model. The train test split was set to 80% for the training set and 20% for the test set.

The followings steps were taken to improve object detection tasks.

- Increasing the input resolution of the network to 608 * 608 to increase the precision.

- Using different resolutions, orientations, brightness, and contrast of the input image as an augmentation technique. Here, the flipping of images is disabled because the flipped traffic signs may denote two different meanings. For example, a flipped version of curve left may belong to the opposite class of curve right.

- Checking whether all the images were correctly labelled and if not label them.

- The anchors were recalculated for the training set using K-means clustering. It increased the precision and mean average precision.

TABLE IV shows the results extracted from the experiments done by authors of Mapillary Traffic Sign dataset [15]. The classifier combined with ResNet101 has reached 83.4 mAP for 313 classes while ResNet50 has reached 81.1 mAP value.

TABLE IV.  Baseline experiments carried out by the authors of Mapillary Traffic Sign Dataset [15]

| Model | mAP |
|---|---|
| FPN50 + classifier | 81.1 |
| FPN101 + classifier | 83.4 |

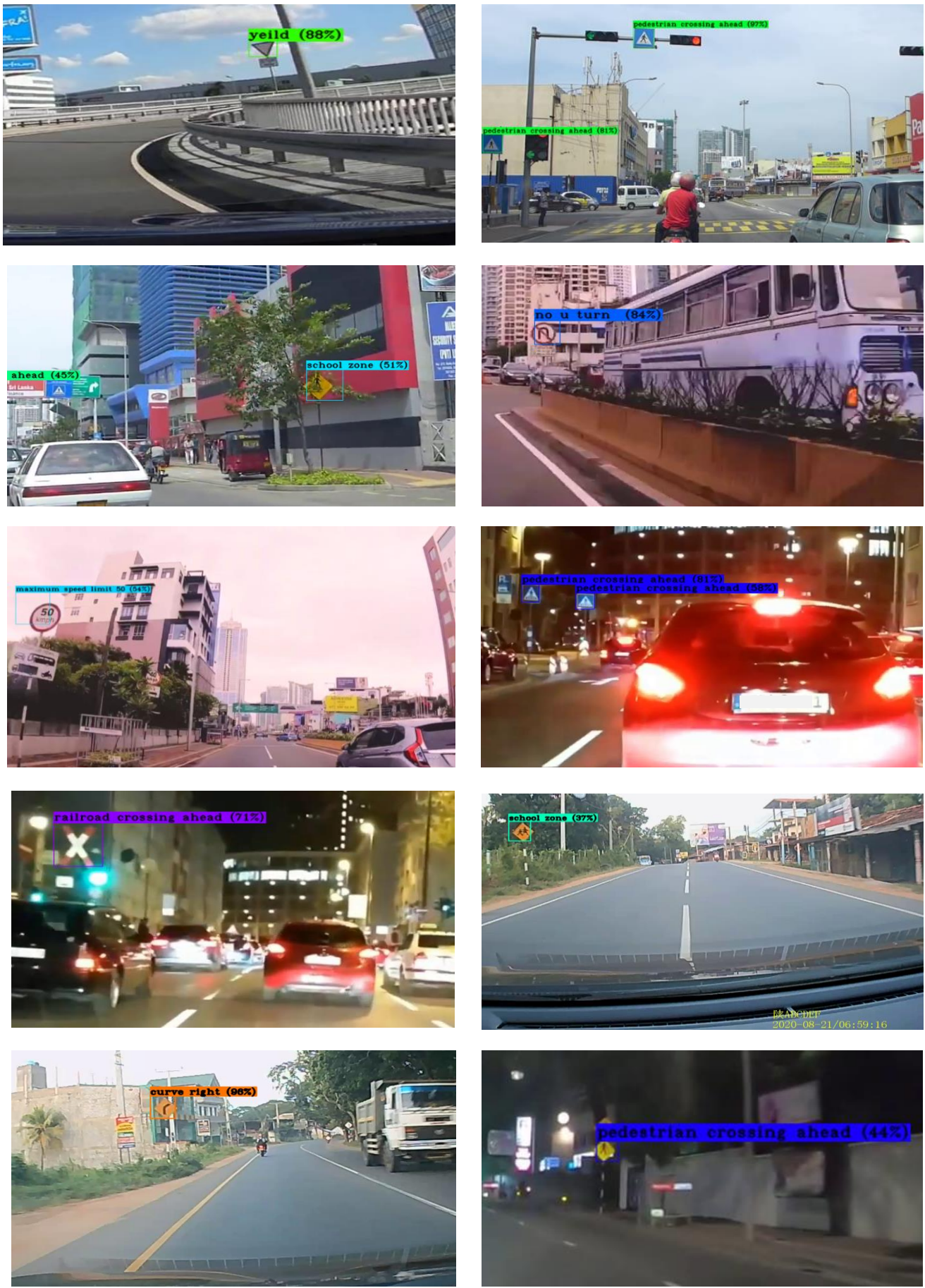

Fig. 7.   Traffic sign detection results obtained on different environmental and lighting conditions.

Under the configurations given in Table I, the proposed system was able to detect objects at an average of 55 FPS with a mean average precision of 64.71%. Some example traffic sign detection results including detections at Sri Lankan roads under different lighting and environmental conditions are shown in Fig. 7.

## V. Conclusion

In this work, we have presented a robust real-time traffic sign detection approach with the detection speed of 55 FPS. We have also achieved a mean average precision of 64.71%. The accuracy of the presented approach can be increased further by adjusting the configurations of the YOLO architecture and tuning the hyper-parameters while maintaining the detection speed consistently. The adverse effect of partially occluded traffic signs, damaged traffic signs, and extreme weather conditions can further be decreased by applying techniques such as presenting the CNN with partly visible signs, applying 3D reconstruction algorithms [16] and fuzzy C-means clustering [17].

The model we presented in this paper can detect traffic signs at a very high frame rate of 55 FPS and could achieve mean average precision of 64.71%. Having a frame rate of over 30 FPS guarantees the real-time performance of the system. Further, the voice assistant feature along with accurate detection can solve most of the problems which are caused due to the missing or not being aware of the traffic signs.

In future, we would like to extend the training of the model to the whole dataset with high-end GPU devices. To increase the accuracy further, we see the potential of modifying the architecture of YOLO detector. With the accuracy increased, we can embed the system into a single board PC within the vehicles to assist the drivers.